# Screening of Pneumonia and Urinary Tract Infection at Triage using TriNet: First Step to Machine Learning Based Medical Directives

Stephen Z. Lu

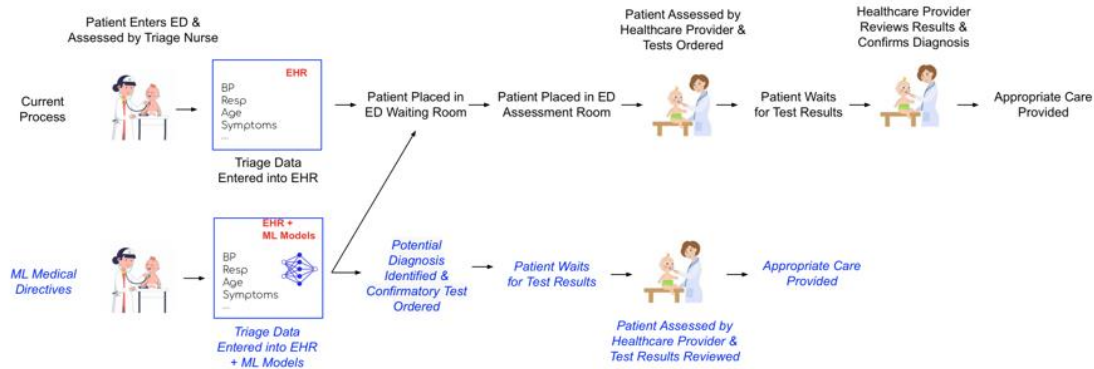

Fig. 1. The current process (top) outlines the baseline ED process where patients are placed in a waiting room and await PIA. The TriNet process (bottom) reduces patient length of stay by ordering appropriate testing prior to PIA according to the model's output.

## Abstract

*Due to the steady rise in population demographics and longevity, emergency department visits are increasing across North America. As more patients visit the emergency department, traditional clinical workflows become overloaded and inefficient, leading to prolonged wait-times and reduced healthcare quality. One of such workflows is the triage medical directive, impeded by limited human workload, inaccurate diagnoses and invasive over-testing. To address this issue, we propose TriNet: a machine learning model for medical directives that automates first-line screening at triage for conditions requiring downstream testing for diagnosis confirmation. To verify screening potential, TriNet was trained on hospital triage data and achieved high positive predictive values in detecting pneumonia (0.86) and urinary tract infection (0.93). These models outperform current clinical benchmarks, indicating that machine-learning medical directives can offer cost-free, non-invasive screening with high specificity for common conditions, reducing the risk of over-testing while increasing emergency department efficiency.*

## 1. Introduction

Since 2014, wait-times in Canadian emergency departments (EDs) have increased over 17%, contributing to delays in crucial patient diagnosis and treatment administration [1]. With the traditional ED workflow outlined in Figure 1, patients who are not critically ill are placed in the waiting room after an initial assessment by a triage nurse. Subsequently, patients are moved into an ED assessment room and await their physician initial assessment (PIA). After the PIA, a confirmatory test is ordered according to the suspected diagnosis, triggering another waiting period for the test results before appropriate care can be delivered. Considering that the length of stay for 90% of patients in Ontario has reached 8.3 hours [2], this traditional ED workflow does not satisfy the increasing needs in modern healthcare.

One widespread strategy for increasing patient flow is the use of nursing medical directives at triage [3], [4]. These directives enable nurses to order specific investigations such as bloodwork, urine testing, and imaging before the patient is seen by a physician. Medical directives are designed to provide the physician with more information prior to their initial assessment, leading to improved decision-making and reduced wait-times for patients overall. As seen in [28], critical to this approach is the high positive predictive value (PPV) of any directive since false positives are indicative of unnecessary (and potentially invasive) testing, while false negatives result in the current standard of care. To minimize false positive testing, medical directives require nurses to interrupt the triage workflow in order to precisely assess if a patient satisfies the inclusion criteria for a specific diagnostic test. Therefore, conventional medical directives are limited by the nurse's workload and experience, preventing its large-scale implementation due to concerns over inaccurate diagnoses and invasive over-testing [8].

We describe here a novel machine learning (ML) model, named TriNet, that offers first-line screening with high PPV for pneumonia and urinary tract infection (UTI). As illustrated in Figure 1, TriNet can be used to automate the medical directive assessment process, relieving nursing workload while increasing the precision and specificity of the screening test. Trained on triage data and past medical information from electronic health records (EHR), TriNet's risk-free screening test is non-invasive, cost-free and instantaneously returns predictions from the hospital's real-time EHR system. Upon positive diagnosis, the appropriate confirmatory test (see Appendix 6.1) can be ordered while the patient is placed in the waiting room and completed prior to PIA, improving physician judgment and productivity while decreasing patient length of stay. If TriNet returns negative, the patient will travel through the typical baseline ED process. Therefore, taking TriNet's two-pronged screening approach at triage increases ED workflow without risking missed diagnoses.

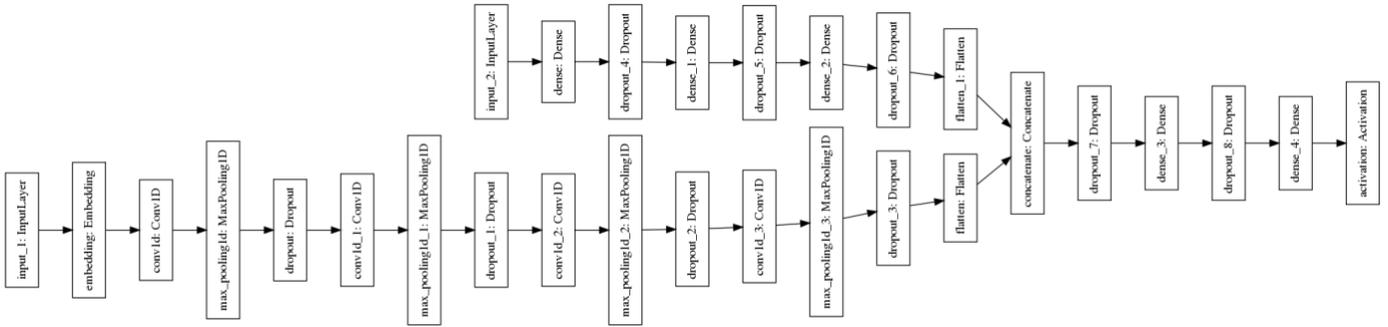

Fig. 2. TriNet is a modular neural network that contains two sub-branch models. The first is a 1D CNN that performs text classification on medical notes and the second is a fully connected feedforward network that analyzes categorical and continuous features. The two branches merge into a concatenated layer and produce a sigmoid activated output layer for binary classification.

## 2. Methods

We hypothesized that TriNet could predict pneumonia and UTI with high PPV in all patients presenting to ED at triage. These two conditions were selected because they often require additional testing (see Appendix 6.1) for clinicians to verify the diagnosis and would greatly benefit from highly precise medical directives. The pool of conditions targetable by TriNet can be expanded to additional presentations if high PPV can be achieved.

### 2.1. Data

De-identified and anonymous data was collected from Epic EHR from a major tertiary care hospital in a large urban city for all patients presenting to ED from July 2018 to June 2019. The total number of authentic positive labels for pneumonia and UTI account for approximately 6% of all patients observed in the ED. Labels for pneumonia and UTI were determined using diagnostic labels selected by doctors from a drop-down menu when completing a patient's EHR file. Model input features are detailed in Appendix 6.2. Missing values for continuous and categorical features were replaced with mean values and most frequent categorical values, respectively. Training (70%), validation (15%), and held out test (15%) sets were generated through stratified sampling to maintain the same ratio of positive to negative samples in each set. To assist with class imbalance, SMOTE up-sampling of the minority class was performed on the training set with target ratio set to 1.0 and 15 k-neighbor estimators. Notably, the validation and test sets did not undergo any permutations.

When a specific diagnosis is suspected, confirmatory clinical testing would typically be ordered. In Table 1, we list the PPV performance of clinicians for diagnosing pneumonia and UTI based on patient history and a physical exam alone, collected from past literature. Considering that both conditions often require additional testing for diagnosis confirmation, note that their benchmark PPVs are particularly low as indicative of a high false positive rate. In TriNet's approach, our goal is to threshold the PPV significantly higher than that achieved by clinicians, thus minimizing the harm and cost of unnecessary testing to justify the automation of downstream testing. Even though this might come at the cost of reduced model sensitivity, there are no pervasive consequences for false negative predictions since patients will undergo the baseline ED process.

### 2.2. Network Architecture

TriNet adopts the architecture of a Modular Neural Network (MNN) that decomposes the classification process into two independent sub-branches in a divide and conquer approach. The first sub-branch consists of a 1D convolutional network (CNN) with an embedding layer that analyses written nursing notes containing information such as past medication, current symptoms and miscellaneous remarks in order to distinguish textual patterns in cases of pneumonia and appendicitis. Using BioSentVec, a pretrained medical embeddings model developed by Chen et al. [15], the notes are converted into word vectors according to semantic similarity. Then, these vectors are fed into TriNet's embedding layer and trained on 4 consecutive layers of ReLU activated 1D convolutions and max pooling. To prevent overfitting, dropout layers were added as well as l2 kernel regularization for the first two convolutional layers.

The second sub-branch of the MNN consists of a simple feedforward network that analyzes the other numerical features of the input vector (see Table 2). Composed of 3 fully connected layers with ReLU activation and dropout, this network merges with the CNN to produce a concatenated dense layer as well as a final output layer adjusted for binary classification activated by a sigmoid function. TriNet was optimized using stochastic gradient descent with a learning rate of 0.00001, a momentum of 0.9 and a weight decay of 1e-7. More importantly, in order to minimize false positive rate and maximize model PPV, class weighting was applied to TriNet, favoring the positive class according to true class ratio.

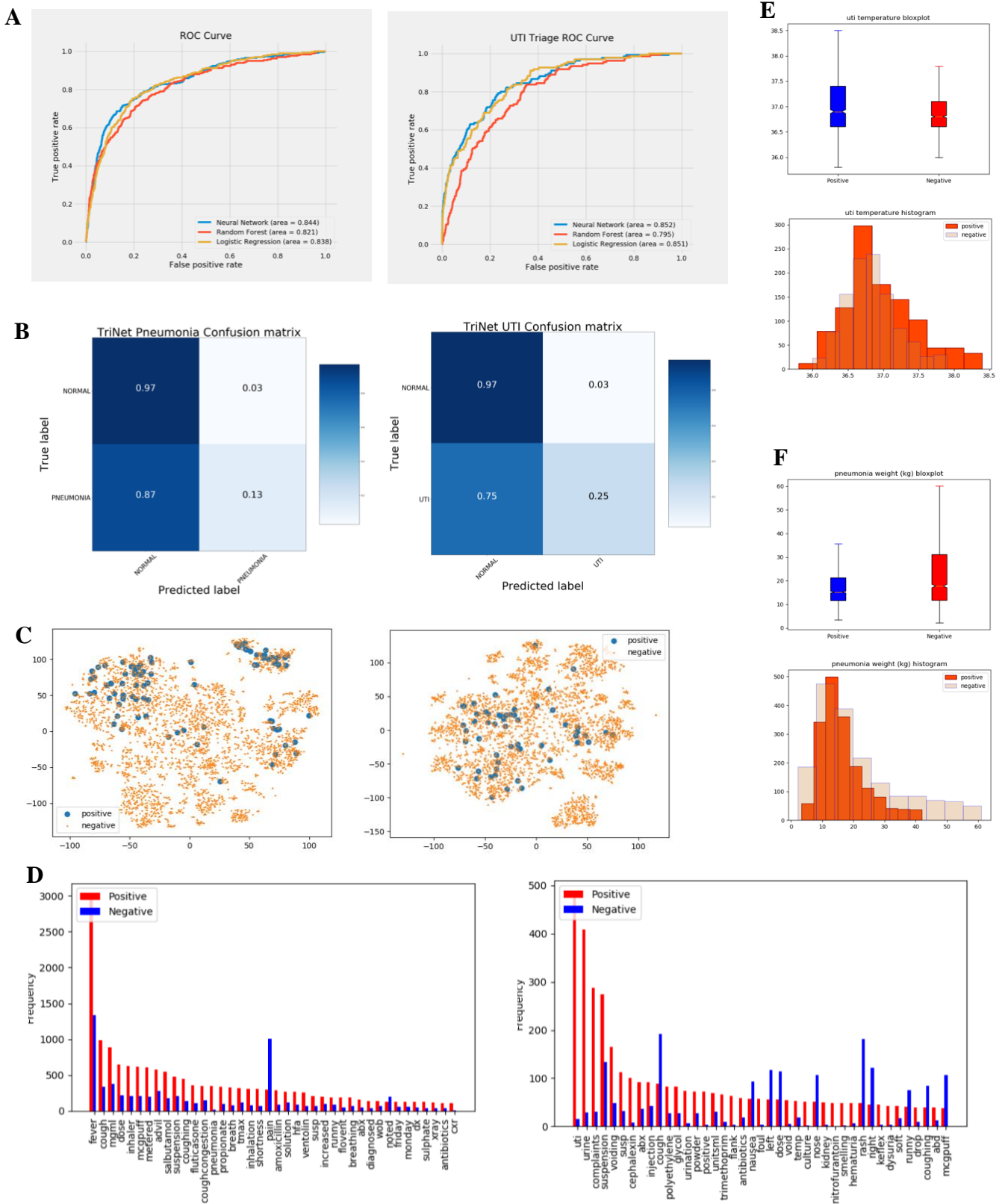

Fig. 3. Training and evaluation plots for pneumonia (left) and urinary tract infection (right). **a** Receiver operating characteristic curves were plotted for pneumonia (left) and UTI (right) predictions; comparing the performance of *TriNet* (blue), logistic regression (yellow) and random forest (red) models. By studying the area under the curve (AUC), we observe that *TriNet* achieves the highest true positive to false positive ratio. **b** Confusion matrices were plotted for *TriNet* after optimizing for maximal PPV. Based on the condition, this method minimizes false positive rate (0.03) while maintaining reasonable sensitivity (0.13~0.25), as is required by clinical models. **c** TSNE dimension reduction was performed on 1D CNN word embeddings to illustrate linear separability in the triage notes. **d** Token frequency is plotted for pneumonia (left) and UTI (right) for words achieving >50% difference in positive to negative frequencies. We observe correlations between word frequency and target condition. **e** Boxplot and histogram are plotted for body temperature in UTI positive and UTI negative patients. **f** Boxplot and histogram are plotted for weight in pneumonia positive and pneumonia negative patients.

## 3. Results

We first compare our approach against documented metrics evaluating clinician performance on patient history and a physical exam alone. We then compare TriNet to recent screening tools using similar data sources to detect community-acquired pneumonia and UTI. Finally, we study the importance of both the 1D CNN and the fully connected branch in achieving high linear separability for our binary classification task.

### 3.1. Evaluation Metrics

TriNet was evaluated and compared according to the following metrics: positive predictive value (PPV), sensitivity (TPR) and specificity (TNR). Confusion matrix, ROC curve as well as training accuracy and loss were plotted for pneumonia and UTI models to further detail TriNet's performance.

| Condition | Method | PPV | TNR | TPR |
|---|---|---|---|---|
| Pneumonia | **TriNet** | **0.86** | **0.97** | 0.13 |
| | Physician [5][6] | 0.27 | 0.84 | 0.74 |
| | Ambreen et al. (2007) [10] | 0.30 | 0.76 | **0.90** |
| | Jones et al. (2012) [11] | 0.51 | - | 0.74 |
| UTI | **TriNet** | **0.93** | **0.97** | 0.25 |
| | Physician [12]-[14] | 0.77 | 0.69 | **0.65** |

Table 1. Our models are compared to physician diagnosis and other screening tools in recent literature. TriNet achieves significantly higher PPV and TNR (specificity) than other methods. This comes at the cost of TPR (sensitivity), but this loss is rarely significant since patients who trigger true or false negative predictions undergo the baseline ED process without suffering any remarkable consequences.

### 3.2. Pneumonia

As illustrated in Table 1, TriNet achieves significantly higher PPV and specificity compared to modern physician diagnosis and recent screening tools. Even though our model has reduced sensitivity (0.13) compared to other methods (0.74~0.90), this loss is essential to the minimal false positive rate (0.03) that TriNet successfully achieves. Moreover, downstream testing for pneumonia diagnosis generally involves obtaining an invasive chest x-ray with low dose radiation, requiring a robust model with few false positives [7]. Using similar feature inputs as recent pneumonia screening tools, TriNet achieves the highest PPV, suggesting that its use in the medical directive process may increase ED workflows without causing invasive over-testing or other harm.

### 3.3. Urinary tract infection

Once more, TriNet achieves higher PPV and specificity than current physician benchmarks for UTI prediction. The downstream test for UTI diagnosis confirmation is urinalysis which requires a urine sample from the patient. This procedure is straightforward and risk-free for adult patients but becomes increasingly challenging for young infants in pediatric setting who often require an invasive urinary catheter to extract the sample. Previous research has shown that urethral catheterization as the initial method for UTI screening is ineffective, costly, painful and time consuming [16]. TriNet's PPV and TPR benchmarks, indicative of minimal false positive testing, suggests that it can effectively enable the automation of urinalysis at triage for ~25% of pediatric UTI patients without risking invasive over-testing.

### 3.4. Embedding Visualizations and Token Frequency

Embedding visualizations were performed on 1D CNN using principal component analysis (PCA) and t-Distributed Stochastic Neighbor Embedding (TSNE) [17] to illustrate word vector relationships and display data linear separability (See Figure 3c). Furthermore, token frequency is charted in Figure 3d for words with significant (>50%) differences between positive and negative samples. We observe significant correlations between these words and the conditions they are attached to, suggesting that the word embeddings are significant in TriNet's ability to achieve high PPV. In Appendix 6.4, a few interesting word definitions are given for pneumonia and UTI conditions.

### 3.5. Feature Importance and Selection

Feature importance is visualized through boxplots and histograms (Figures 3e-3f) comparing positive and negative sample vital sign data. Notably, there is a significant increase in body temperature in UTI positive patients and a significant decrease in body weight in pneumonia positive patients. These results correlate with previous findings [18]-[20] which confirm that UTI patients often develop fevers and that pneumonia is especially prevalent and dangerous in children under the age of 5 years old. These metrics validate TriNet's learning capacity and confirm the relevance of its input features.

## 4. Limitations and Discussion

There is a growing number of patients presenting to the ED. This increase in patient volume contributes to rising patient wait-times, delays in diagnosis, and logistical challenges for healthcare administrators and ED clinicians. We demonstrate an opportunity to leverage ML medical directives with high PPV in a cheap and risk-free screening test to assist with managing large patient volumes. Due to concerns of invasive over-testing resulting from high false positive rates and human practice error, conventional nursing medical directives are limited. TriNet demonstrates that it is possible to achieve moderate sensitivity for identifying common ED presenting conditions while maintaining PPV well above clinical benchmarks of front-line clinicians. Notably, our positive labels for patients seen in the emergency department excludes patients where confirmatory testing was ordered but was negative. Once these are taken into account, we believe that ~25-30% of the patients seen in pediatric EDs would benefit from ML medical directives.

By avoiding giving the clinician any putative diagnosis, our approach also reduces the risk of the clinician relying on TriNet as a replacement of their own judgement. Nevertheless, there still exists a risk that healthcare providers may assume that a patient does not have a specific condition when TriNet has not been triggered and appropriate testing was not pre-ordered. This may contribute to downstream under-testing of a patient with adverse outcomes to their patient care. Therefore, physicians working with TriNet must strictly adopt a non-reliant method of care that ignores the existence of such a screening tool. Investing in human factors engineering along with clinical staff to understand how to work in tandem with ML medical directives will be essential for systems like this to be successfully deployed without adverse impact to patient care.

## 5. Conclusion

TriNet obtained favourable AUROC and PPV for predicting pneumonia and UTI at triage and may enable the automation of downstream testing for subsets of patients adding efficiency to the ED. Further statistical validation and in-situ prospective clinical testing is required and will be the focus of our future work.

## Acknowledgement

We thank Dr. Devin Singh (Hospital for Sick Children) for providing de-identified and anonymous data from a major tertiary pediatric hospital.

# 6.    Appendix

## 6.1. Downstream Testing

Downstream tests are often required by ED physicians to confirm initial diagnosis from PIA. These tests vary greatly in terms of required time, invasiveness and complexity. Therefore, test ordering must be done with care and precision in order to minimize false positive testing and optimize hospital resources, especially when the test is somewhat invasive or time-consuming. In the cases of pneumonia and urinary tract infection, both conditions require minimally invasive tests to confirm PIA diagnosis. Therefore, TriNet's ability to automate the test ordering process for these conditions with minimal false positive predictions suggests that ML based medical directives may effectively expedite ED triage workflow without any pervasive side-effects.

| Condition | Downstream Test | Invasiveness |
|---|---|---|
| Pneumonia | Chest X-Ray | Minimal (Radiation) [21] |
| UTI | Urinalysis | None to minimal (May require catheterization) [20] |

## 6.2. Model Features

To achieve TriNet's benchmarks, our model's feature inputs were carefully selected according to medical insights to maximize linear separability of our classification tasks. These features include textual data that feed into TriNet's 1D conv branch and quantitative/qualitative data that serve as input to the fully connected branch of our model.

| Feature | Description |
|---|---|
| Triage Notes | Medical notes written by triage nurse into EPIC |
| Medication | Past medication reported on patient EHR |
| Symptoms | Presenting symptoms including chief complaint |
| Age | Patient age in months |
| Weight | Patient weight in kg |
| Systolic BP | Systolic blood pressure in mm Hg |
| Diastolic BP | Diastolic blood pressure in mm Hg |
| Respiratory Rate | Number of breaths per minute |
| Temperature | Body temperature in °Celsius |
| Pulse | Patient pulse in beats per minute |
| CTAS | Canadian Triage Acuity Score (1-5) |
| Arrival Method | Patient method of arrival to ED |
| Arrival Time | Time of arrival (hour, weekday, month) |
| Gender | Patient gender (U, M, F) |

## 6.3. Model Training

TriNet was trained on ~1000 epochs and illustrated strong learning curves. Training and validation accuracy and losses are plotted below for pneumonia (top) and urinary tract infection (bottom)

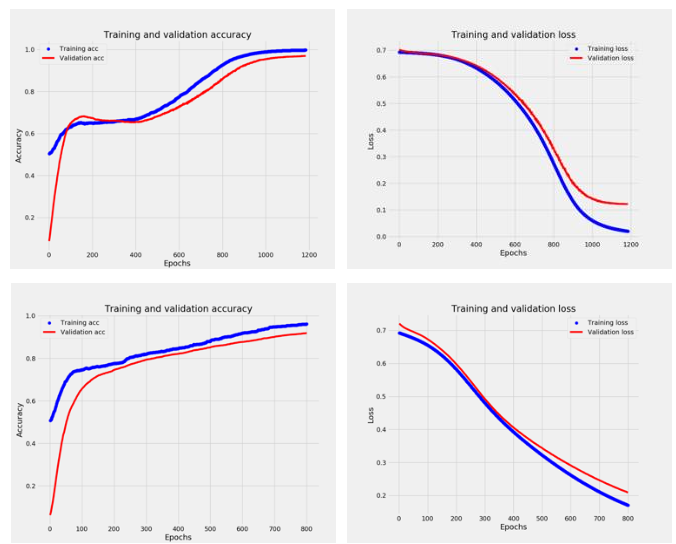

## 6.4. Token Frequency

The following words were selected from Figure 3d to illustrate word frequency correlation with pneumonia and UTI positive patients

| Condition | Word Token | Definition / Correlation |
|---|---|---|
| Pneumonia | Salbutamol | Pneumonia treatment [22] |
|  | Fluticasone | Increased risk [23] |
|  | Amoxicillin | Common treatment [24] |
| UTI | Cephalexin | Antibiotic [25] |
|  | Polyethylene glycol | Laxative [26] |
|  | Trimethoprim | Antibiotic treatment [27] |

## 6.5. Feature Importance

Boxplots and histograms were plotted for age, weight and all vital sign data.

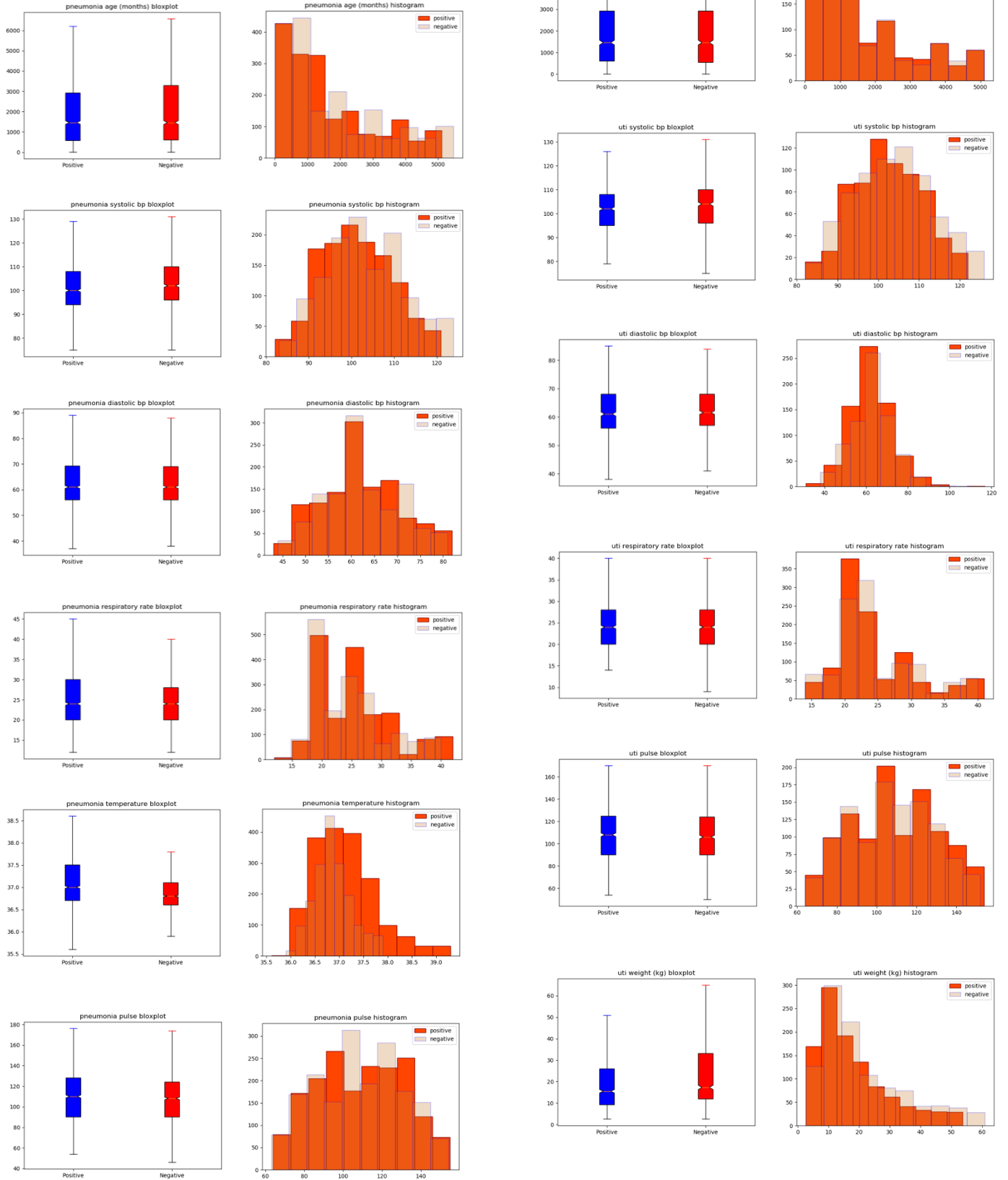